# i-SAIRAS 2024

# UWB Anchor Based Localization of a Planetary Rover

[1]*Nüchter, Andreas, [2]Werner, Lennart, [1]Hesse, Martin, [3]Borrmann, Dorit, [1]Walter, Thomas, [1]Montenegro Sergio & [4]Grömer, Gernot

[1] andreas.nuechter@uni-wuerzburg.de, Julius Maximilian University Würzburg, Germany

[2] ETH Zürich, Switzerland

[3] Technical University of Applied Sciences Würzburg-Schweinfurt, Germany

[4] Austrian Space Forum, Innsbruck, Austria

**Abstract**

Localization of an autonomous mobile robot during planetary exploration is challenging due to the unknown terrain, the difficult lighting conditions and the lack of any global reference such as satellite navigation systems. We present a novel approach for robot localization based on ultra-wideband (UWB) technology. The robot sets up its own reference coordinate system by distributing UWB anchor nodes in the environment via a rocket-propelled launcher system. This allows the creation of a localization space in which UWB measurements are employed to supplement traditional SLAM-based techniques. The system was developed for our involvement in the ESA-ESRIC challenge 2021 and the AMADEE-24, an analog Mars simulation in Armenia by the Austrian Space Forum (ÖWF).

**Introduction and Background**

Many planetary robotics applications require (semi-)autonomous rover operation for which localization is essential [1]. This paper presents the experience gained during the ESA-ESRIC Space Resources Challenge in 2021 and AMADEE-24.

Within the ESA-ESRIC challenge rovers need to traverse a lunar-like terrain, teleoperated through a 6 seconds round trip time delayed network. Semi autonomous operation is thus required. The simulated lighting and terrain conditions resemble a landing spot in the polar regions of the moon, rendering traditional camera based localization error-prone due to blinding and strong shadows. Long shadows from lunar rocks or the rover itself yield high contrast images with moving features, which are less than ideal for camera-based SLAM (simultaneous localization and mapping). Besides the scientific analysis of lunar rocks, a central element of the ESA-ESRIC challenge was to provide a detailed map of the lunar terrain.

Similarly, the AMADEE-24 mission, led by the Austrian Space Forum (Österreichisches Weltraum Forum, ÖWF), was an analog space research field campaign set for 2024 in collaboration with the Armenian Aerospace Agency. The mission emulated the conditions and challenges that astronauts might face on Mars, taking place in a remote location on Earth that closely resembles the Martian environment. AMADEE mission sites are typically chosen for their similarity to the Martian landscape, featuring rocky terrains, extreme temperatures, and minimal vegetation. Previous AMADEE missions have been conducted in deserts like in Oman and Israel, offering representative analog conditions. The location for AMADEE-24 in a desert in Armenia has been selected to provide a challenging environment that closely resembles the conditions astronauts might encounter on the Red Planet, cf. Fig. 1.

The primary mission goals include testing new technologies, refining operational procedures, and studying human factors in space exploration. AMADEE-24 evaluated spacesuits, rovers, and habitat systems while simulating daily operations like extravehicular activities and habitat maintenance. It also investigates the psychological and physiological effects of isolation and confinement on the crew.

To both missions, we contribute a rover system equipped with a 3D LiDAR scanner for mapping. For aligning the 3D laser scans to get a global map of the environment, preliminary pose estimations needed to be attached to the individual scans [4]. We use a move-and-wait scheme, where the operator decided on a waypoint based on a 3D laser scan and used several RGB camera images for situational awareness. Sharing a single destination pose accounts for the communication constraints present in planetary missions. For the field operations, a graphical user interface is implemented to allow for easy selection of the next waypoint [12]. The rover then drives autonomously to the target destination, after which the cycle is repeated. This driving mode required knowing the robot pose at all times, so localization wass crucial.

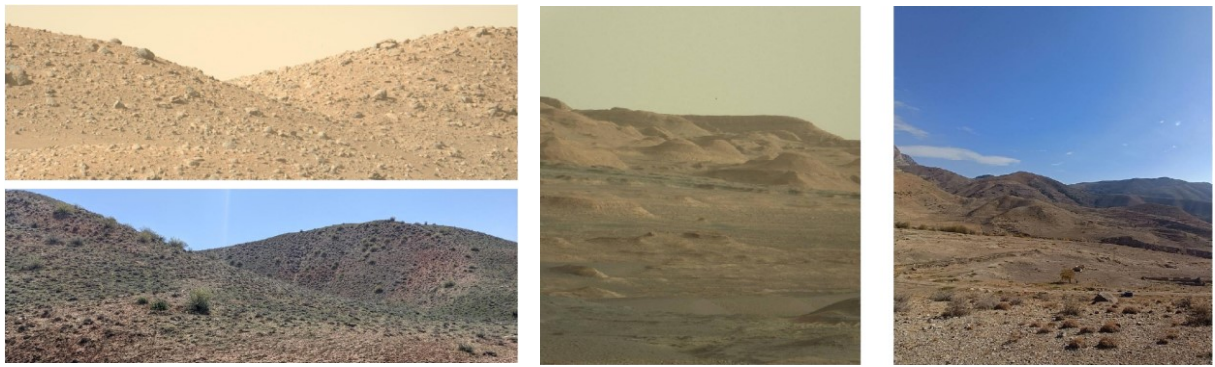

Figure 1: Left: A noteworthy similarity lies in the convergence patterns of the slopes, indicating a consistent geomorphic process at play in both locations. Additionally, the comparable steepness of the hillsides suggests uniform geological influences shaping these features, which offer insights into the dynamic forces shaping landscapes on Mars and Earth. (Mars photo: NASA/Perserverance Rover, Image taken west to Belva Crater, Mars Region Jezero Crater Longitude: 77.36869069° Latitude:18.48280163° (Sol 784). Middle and right: The shared features suggest a comparable geological context, which implies the similar underlying processes shaping these mountainous terrains. (Mars photo: NASA Curiosity Rover, Mars Region Gale Crater, Base of Mount Sharp, Longitude: 137.36913767° Latitude: -4.673087126129127 ° (Sol 1144)).

RGB-D cameras and wheel odometry are readily available and often used methods for relative self-localization of mobile robots but they are prone to errors that are common in the given scenarios and lead to loss of localization if they occur. RGB-D cameras are known for their high update rate, precision and onboard data processing, but they are sensitive to changing lighting conditions. For a short period of time, wheel based odometry provides a good high frequency pose estimation for the drive controller [5] but wheel slippage on sandy or rocky ground deteriorates those measurements.

Therefore, we suggest a robust technology whereby a mobile system initially disseminates a set of Ultra Wideband (UWB) transceivers to establish a location-aware wireless sensor network (WSN). This enables the robots' location to be determined in a globally stable coordinate system through trilateration. While using a combination of visual and wheel odometry for relative localization, the robot updates its pose regularly based on range measurements to these UWB anchors.

**State of the Art**

Simultaneous Localization and Mapping (SLAM) is a core technology for robotic planetary exploration, enabling autonomous systems to navigate unknown environments while incrementally building a map and localizing within it. The challenges posed by planetary exploration—such as limited communication, harsh environmental conditions, and the absence of GPS—require advanced and robust SLAM solutions.

Visual-based SLAM, using cameras for localization and mapping, has gained prominence in planetary exploration. Techniques like ORB-SLAM [2] and DVO-SLAM [3] exploit visual features and direct methods to construct detailed maps. The Mars rovers (e.g., Perseverance and Curiosity) leverage stereo vision and monocular cameras for mapping and localization. Lidar sensors provide precise range measurements, making Lidar SLAM highly effective for planetary exploration. Algorithms like Cartographer [6] and LOAM (Lidar Odometry and Mapping) [7] have demonstrated the ability to generate accurate 3D maps. A lidar's resilience to lighting conditions makes it suitable for the dim or dusty environments typical of planetary surfaces. Fusing data from multiple sensors (e.g., cameras, Lidar, IMUs) enhances SLAM robustness. This fusion approach is crucial for planetary exploration, where sensors might fail or provide incomplete data. Extended Kalman Filters (EKF) and Factor Graphs are widely used for current state of the art SLAM systems [8]. However, robust loop closure detection, where the robot recognizes or revisits previous locations, is essential for long-term autonomy.

Due to its low cost Ultra-Wideband (UWB) technology has gained significant attention in recent years for localization applications. UWB operates over a wide frequency range (3.1 to 10.6 GHz) and is known for its ability to provide highly accurate ranging and positioning, even in challenging environments. UWB-based localization is widely used in applications such as indoor positioning, industrial automation, and autonomous systems [15], [16]. UWB localization is primarily based on Time of Flight (ToF), Time Difference of Arrival (TdoA), or phase measurements for angle of arrival (AOA) measurements. These techniques calculate the distance between devices by measuring the time taken for a UWB signal to travel between them. TDoA-based systems are particularly effective in multi-anchor setups, providing centimeter-level accuracy. Two-Way Ranging (TWR) is a widely used method in UWB localization, where a device exchanges UWB signals with reference anchors to determine its position [9]. In a typical setup, two devices (tag and anchor) send signals back and forth, and the system measures the round-trip time to calculate the distance. TWR is effective in reducing synchronization issues, making it robust for real-time localization. Given the determined distances to several known UWB anchors the location of a mobile system equipped with an UWB device is calculated using trilateration. In the context of planetary surface exploration, our system operates without anchors deployed prior to an excursion, but the anchors are distributed by the mobile system itself.

Some current implementations for mobile robot localization are Bluetooth-based, due to the widespread availability of Bluetooth Low Energy (BLE) devices, making it a cost-effective solution for positioning and real-time tracking. However, the algorithms use tri- or multilateration based on Received Signal Strength Indicator (RSSI), measuring the strength of the Bluetooth signal, and the distance between a Bluetooth beacon (or anchor) and the target device is estimated based on the inverse square law of signal propagation, which is usually less accurate than ToF measurements.

In previous work, we have developed several SLAM algorithms, mainly using Lidar systems [4, 6] but also integrating Visual SLAM and IMU measurements [13]. However, UWB localization offers significant advantages in complementing SLAM algorithms, particularly in environments where visual and lidar-based systems are limited. By providing robust, accurate, and absolute positioning data, UWB can mitigate the challenges of feature-sparse environments, and adverse conditions as in the polar regions (low sun incidence angles), improving reliable localization and mapping in planetary exploration.

**Approach**

We developped a navigation solution for augmenting camera and wheel based robot odometry in low visibility and high slippage environments. The position of the rover is determined simultaneously by range measurements to previously deployed UWB anchors. Our approach generates a smooth and reliable trajectory for a planetary rover equipped with a 3D Lidar to provide an initial pose estimate for registration of the 3D point clouds.

The operator chooses a next goal pose that the drive controller of the rover approaches using the current pose as feedback. As a main source of relative localization the rover uses an Intel Realsense T265 stereo camera. Its high update rate, precision and onboard data processing capacity makes it a good choice for experimental use cases. However, in challenging environments such as planetary surroundings, the pose determination is degenerated or denied. In those cases, navigation is switched to simple wheel based odometry. For a short period of time, wheel based odometry can provide a high frequency pose estimation for the drive controller [5]. After the destination pose is reached according to visual or wheel odometry, a discrepancy accumulates between the estimated and the actual pose. To reset this error, the position of the rover is determined simultaneously by range measurements to previously deployed UWB anchors. Upon command by the operator, the current odometry pose is overwritten with the UWB position. This operation keeps the position information globally stable as needed by the previously mentioned alignment of the 3D point clouds of the 3D-Lidar scanner and the next position can be approached using relative localization again.

**Implementation**

For the UWB ranging, we use the DecaWave (now Qorvo) DWM1000 transceivers. They are used within a PCB stack originally designed as a flight controller for small UAVs. The DWM1000 interfaces with a STM32F407VG Micro-processor, which handles all data processing and communication to the host computer, cf. Figure 2. Our anchors are battery powered using 9V Lithium blocks.

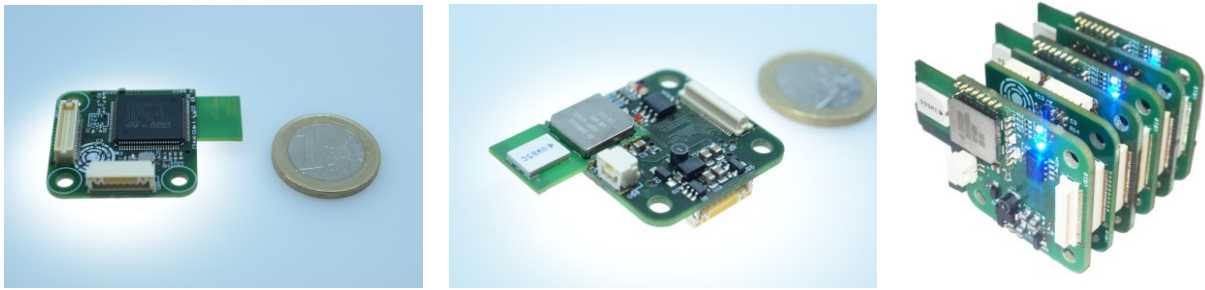

Figure 2: Left and middle: The SKITH-STM32 board, including the sticking out UWB-Tranceiver, comparing to a Euro coin. Right: Stacked SKITH-boards.

The hardware shown in Figure 2 has a modular structure. Each module consists of a circuit board with the dimensions 30.5mm x 30.5mm, the board thickness is 1.6mm. The modules can be stacked at a distance of 5mm using board-to-board connectors with 50 electrical contacts. For mechanical fastening, each module has three holes with a diameter of 3.2mm.

The voltage regulator implemented on this SKITH-STM32 (STM32F4, UWB transceiver) board has an input voltage range of input voltage range from 4.4V to 50V and delivers an output voltage output voltage of 3.3V at a maximum current of 500mA. This voltage is applied directly to the contacts of the board-to-board connector. It is also used to power the ICs on the board. These include the STM32F407 microcontroller, a UWB radio transceiver DWM1000, and a CAN transceiver. This module can be regarded as a base board. Its functionality can be extended by connecting one or more of the following modules. It is also possible to connect several SKITH-STM32 modules in order to multiply the computing power. In this case, communication between the microcontrollers can take place via CAN-bus. The interfaces that are led out via the board-to-board connector include UART, SPI, CAN, I2C, ADC, PWM, GPIO. There exist SKITH-boards with (1) STM32F4 and UWB transceiver, (2) H-bridge and stepper motor driver, (3) USB-to-serial, 3D accelerometer, gyro (LSM9DS1) and magnetic field (BMP388) and servo motor and SD-card interfaces (4) GNSS receiver (ublox SAM-M8Q) and wireless LAN (ESP32) as well as (5) a prototyping board, cf. Figure 2.

One UWB node is directly attached to the rover. The microcontrollers of the UWB boards run the operating system RODOS [10]. With the RODOS-to-ROS bridge this UWB node is connected to the Robot Operating System (ROS) [11] middleware running on the linux-based central computer of the rovers.

Initially, the UWB anchors have to be distributed over the drive area. To not waste mission time and risk damaging the robot, a positioning at distance is needed. A substantial separation of about 15m needs to be reached with the deployment mechanism to cover the area needed for the ESA-ESRIC challenge. A similar area is covered in front of the habitat during AMADEE-24. The long distance deployer is essentially a compact 3D-printed, $CO_2$ powered rocket with a DWM1000 UWB transceiver board as payload. For propulsion, a 15g $CO_2$ cartridge as commonly used for beer dispensers is mounted on the rocket. At launch, a spring-loaded striker opens the cartridge within an enclosed launcher tube, propelling the anchor rocket. The two launcher tubes with attached striker mechanisms are shown in Figure 3. That allows to span a coordinate system by launching the two anchors at an angle of approximately 90 degrees.

The origin anchor on the other hand is dropped off the delivering rover. A simple mechanism with a preloaded spring and pin-pulling servo ejects the anchor reliably. The deployer and anchor capsule are depicted in Figure 4.

The Ultra Wideband (UWB) transceivers perform symmetrical double-sided two-way ranging (SDS-TWR), a ranging method that employs two delays inherent to signal transmission to ascertain the distance between two stations, thereby obviating the necessity for clock synchronization between the UWB nodes. We used a similar setup to [9] , where also the Double Sided Two Way Ranging is described in greater detail. The position of the rover is determined simultaneously by trilateration using range measurements to previously deployed UWB anchors.

During the mission phase, a coordinate transformation must be found such that the pose of the robot in the camera odometry frame can be determined by an UWB measurement. Only this transformation allows for a seamless swap from camera odometry to UWB enhanced wheel odometry. Once the anchor nodes have been distributed, a calibration drive is perfomed to align the coordinate systems of the UWB localization and the robot using the Intel T265 stereo camera. Due to the higher frequency of the visual odometry, a pair of positions is stored for subsequent processing whenever a new UWB position is available during the calibration run. To obtain the transformation, the problem is formulated as a non-linear least squares problem and solved for a 2D transformation that minimizes the distances between the corresponding positions with the Ceres Solver [14]. Once the coordinate frames of the UWB localization system have been aligned with the visual odometry of the mobile robot, which is achieved through the use of an Intel T265 stereo camera, the robot obtains a globally stable coordinate system for localization.

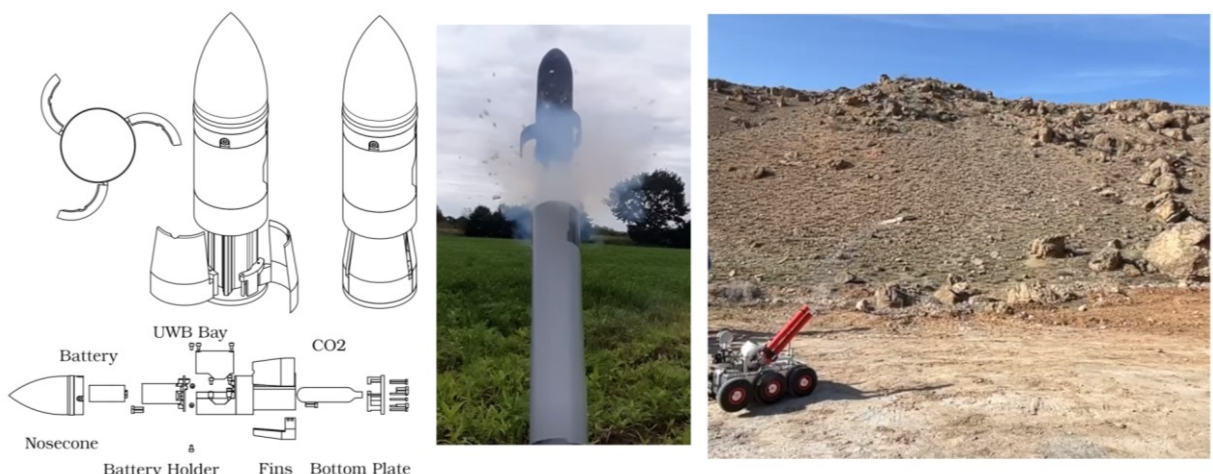

Figure 3: Left above: Anchor in flying (l) and stored (r) configuration. Left below: Composition of the anchors. Middle and right: Launch of the deployers.

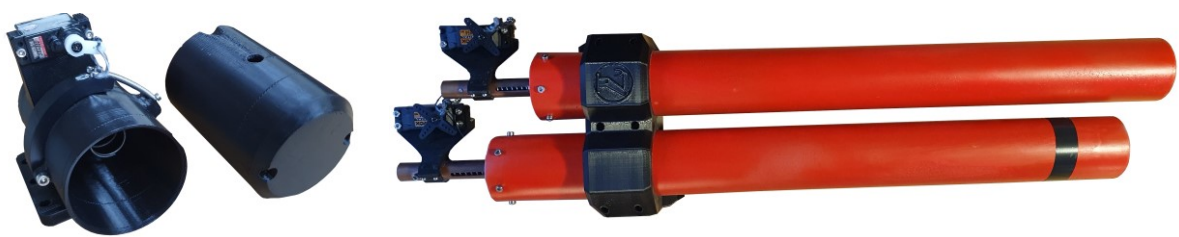

Figure 4: Left: Origin Deployer with Anchor Capsule. Right: Launcher Tube Assembly with Strikers.

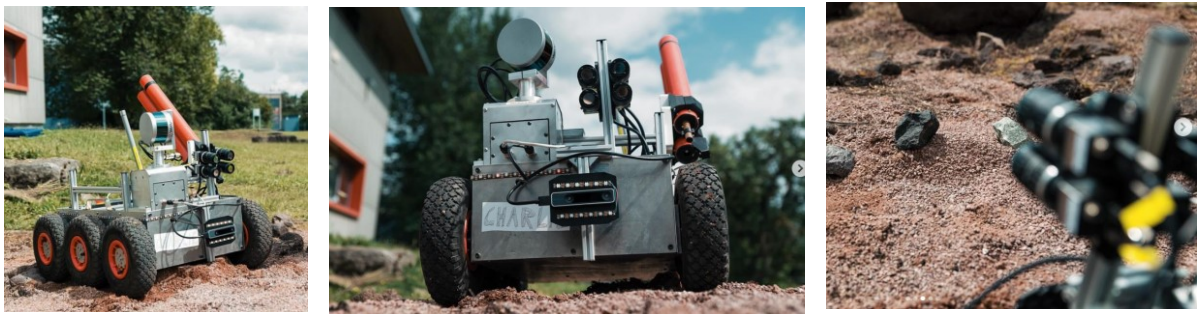
Figure 5: Left and middle: The autonomous mobile robot Charlie. Right: geological analyses.

**Experiments and Results**

The autonomous mobile robot Charlie is skid-sterred and powered by two 90W Maxon motors, cf. Figure 5. Its sensor suite includes an Intel Realsense T265, a Velodyne PUCK laser scanner that is mounted vertically and spins around the up-axis and 4 industrial IDS cameras with different filters, which are used for geological experiments, i.e., for the characterization of rocks. One UWB tag is attached to the back of the rover.

For a quantitative analysis of the range measurement we perform in-lab experiments. We analyze the accuracy of the range measurements by comparing the given values to a genuine truth, we obtained by a ruler. Figure 6 shows typical results, where the ranging inaccuracy is about 40 to 70 centimeter. The subsequent processing algorithms must handle this imprecision.

Given the TWR measurements and collecting these at the robot, we perform triangulation. Fig. 6 shows an anchor distribution as ground truth and the resulting reconstructed positions. Also here, the values are overestimated. For this study, we calculate in 2D and use a single node on the robot. By using more than one receiver at the mobile robot, the localization becomes more stable and estimating of orientation becomes possible. Nevertheless, we currently utilize planar calculations to reduce complexity. Reliable 3D measurements require a significant change in altitude for one anchor which cannot be guarantueed in a general setup. Additionally, wheel odometry is also restricted to planar poses and the expected improvement in accuracy is therefore expected to me minor.

**Conclusions**

This paper motivates the usage of UWB-based localization using TWR in planetary exploration. We have presented a mobile robot system, that is capable of distributing up to 5 sensor nodes in an environment before exploring it. The system has been successfully applied during the AMADEE-24 mission (cf. Figure 3 (right)). Evaluations show that the localization provides sufficient accuracy to complement SLAM methods. Future work will integrate these results in an Extended Kalman Filter framework for robotic mapping.

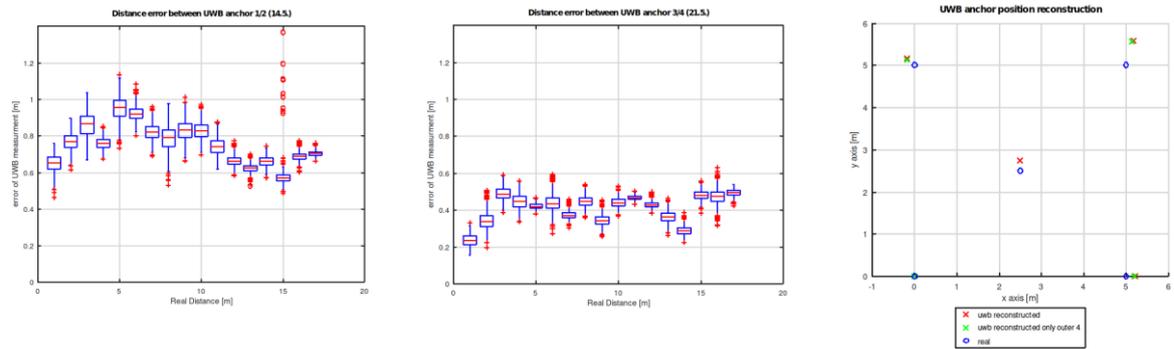

Figure 6: Left and middle: Range measurements for different anchors in an indoor setup. The actual range is often overestimated and shows a large variance. Right: Result of a triangulation. Ground truth vs. the triangulation results using all 5 resp. the 4 outer nodes are given.